\documentclass[journal]{IEEEtran}
\pdfoutput=1
\usepackage{subfigure}
\usepackage{blindtext}
\usepackage{graphicx}
\usepackage{cite}
\usepackage{csquotes}
\usepackage{caption}
\usepackage{subfig}
\usepackage{amssymb}
\usepackage{amsmath}
\usepackage{siunitx}
\usepackage{multicol}

\usepackage{xcolor,colortbl}
\definecolor{Gray}{gray}{0.90}
\newcolumntype{a}{>{\columncolor{Gray}}c}
\usepackage{xcolor,soul}
\definecolor{light-gray}{gray}{0.95}
\sethlcolor{light-gray}

\graphicspath{{Images/}}
\newcommand{\fig}[1]{Fig.~\ref{#1}}
\newcommand{\tb}[1]{Table~\ref{#1}}
\newcommand{\eq}[1]{(\ref{#1})}
\DeclareCaptionLabelSeparator{periodspace}{.\quad}
\usepackage{enumitem,booktabs}
\usepackage[referable]{threeparttablex}
\renewlist{tablenotes}{enumerate}{1}
\makeatletter
\setlist[tablenotes]{label=\tnote{\alph*},ref=\alph*,itemsep=\z@,topsep=\z@skip,partopsep=\z@skip,parsep=\z@,itemindent=\z@,labelsep=.2em,leftmargin=*,align=left,before={\footnotesize}}
\makeatother

\begin{document}
\title{Neuromorphic Architecture for the Hierarchical Temporal Memory}

\author{Abdullah M. Zyarah,~\IEEEmembership{Student Member,~IEEE,}
        Dhireesha Kudithipudi, ~\IEEEmembership{Senior Member,~IEEE,} \\ Neuromorphic AI Laboratory, Rochester Institute of Technology
\thanks{Manuscript received August 2, 2017; revised March 23, 2018; accepted June 9, 2018. This work was supported by the Seagate.
(Corresponding author:Dhireesha Kudithipudi.)

Abdullah M. Zyarah and Dhireesha Kudithipudi are with Department of Computer Engineering, Rochester Institute of Technology, NY 14623 USA (e-mail: amz6011@rit.edu, dxkeec@rit.edu).

Digital Object Identifier 10.1109/TETCI.2018.2850314
}}

\IEEEoverridecommandlockouts
\IEEEpubid{\makebox[\columnwidth]{~XXXX-XXXX
\copyright2018
IEEE \hfill} \hspace{\columnsep}\makebox[\columnwidth]{ }} 

\markboth{IEEE TRANSACTIONS ON EMERGING TOPICS IN COMPUTATIONAL INTELLIGENCE}{Shell \MakeLowercase{\textit{et al.}}: A Novel Tin Can Link}

\maketitle

\begin{abstract}
A biomimetic machine intelligence algorithm, that holds promise in creating invariant representations of spatiotemporal input streams is the hierarchical temporal memory (HTM). This unsupervised online algorithm has been demonstrated on several machine-learning tasks, including anomaly detection. Significant effort has been made in formalizing and applying the HTM algorithm to different classes of problems. There are few early explorations of the HTM hardware architecture, especially for the earlier version of the spatial pooler of HTM algorithm. In this article, we present a full-scale  HTM architecture for both spatial pooler and temporal memory. Synthetic synapse design is proposed to address the potential and dynamic interconnections occurring during learning. The architecture is interweaved with parallel cells and columns that enable high processing speed for the HTM. The proposed architecture is verified for two different datasets: MNIST and the European number plate font (EUNF), with and without the presence of noise. The spatial pooler architecture is synthesized on Xilinx ZYNQ-7, with 91.16\% classification accuracy for MNIST and 90\% accuracy for EUNF, with noise. For the temporal memory sequence prediction, first and second order predictions are observed for a 5-number long sequence generated from EUNF dataset and 95\% accuracy is obtained. Moreover, the proposed hardware architecture offers $1364X$ speedup over the software realization. These results indicate that the proposed architecture can serve as a digital core to build the HTM in hardware and eventually as a standalone self-learning system.
\end{abstract}

\IEEEpeerreviewmaketitle

\section{Introduction}
\IEEEPARstart{H}{ierarchical} learning is the natural way for biological systems to process sensory inputs, learn information, and accumulate knowledge for goal-oriented behaviors. Hierarchical temporal memory (HTM) is a model inspired by the memory-prediction principle of the brain, and builds its foundation on the hierarchical, structural and information processing properties of the neocortex~\cite{on_intelligence,george_htm}. To design biologically plausible intelligent information processing systems for embedded and energy-constrained platforms, it is imperative to realize efficient hardware models of HTM. These systems have applications in several domains including images recognition and classification~\cite{HTM_Zeta1_HW_7,melis2009study, xing2012bio}, prediction~\cite{EU_Numbers}, natural language processing, and anomaly detection~\cite{anamoly_whitepaper25,lavin2015evaluating}. At a higher abstraction, HTM is basically a memory based system that can be trained on a sequence of events that vary over time. In the algorithmic model, this is achieved using two core units, spatial pooler (SP) and temporal memory (TM), called cortical learning algorithm (CLA). The SP is responsible for transforming the input data into sparse distributed representation (SDR) with fixed sparsity, whereas the TM learns sequences and makes predictions~\cite{cla_whitepaper}. 

A few research groups have implemented the first generation Bayesian HTM. Kenneth et al., in 2007, implemented the HTM network on an FPGA platform targeting image content recognition applications~\cite{HTM_Zeta1_HW_7}. The network is composed of 81 computational nodes arranged in 3 hierarchical layers and offers 148x speedup over its software counterpart. In 2013, Pavan et al. proposed a Verilog implementation of the single fundamental unit in HTM, referred to as a node~\cite{HTM_Zeta1_HW_10}. The proposed node architecture can be used to realize the SP in HTM. Unfortunately, this work is not verified for any spatiotemporal tasks and also it is not feasible to apply for the CLA HTM. In 2015, our group proposed SP architecture for 100 columns~ \cite{zyarah2015reconfigurable}. The proposed design is integrated with the support vector machine (SVM) and verified for classification applications. We also proposed early temporal memory designs for prediction in the same year. In 2016, nonvolatile memory based SP implementation is presented by Streat et al. \cite{streat2016non}. This work, however, investigated only the spatial aspect of the HTM considering the physical constraints of commodity NVRAM. Later on, an analog circuit implementation of HTM SP is proposed by James et al. \cite{james2017htm}. Although the proposed design is power efficient, it lacks reconfigurability which is extremely important for learning and making predictions. Weifu Li et al. \cite{li2016hardware}, proposed a full architecture of the HTM algorithm including both spatial and temporal aspects in 2016. The proposed design has 400 columns (2 cells in each column) connected in point-to-point format to the HTM input space, which eventually causes the column to be in active mode even when there is insignificant activity (noise) in the input space. Furthermore, this implementation does not account for potential synapses in the design. To the best of our knowledge, there are no custom large scale hardware architectures in the literature, that demonstrate the full cortical learning algorithm based HTM.

In this work, we propose a reconfigurable and scalable hardware architecture for the HTM core, which includes both the SP and TM. The SP is modeled with identical processing units (known as columns) that create unique sparse representations of the sensory input. The TM learns sequences and makes prediction using identical cells embedded within the HTM columns. The proposed design is validated with two different datasets: MNIST \cite{MNIST} and EUNF (numbers)~\cite{EU_Numbers}. This neuromorphic architecture can serve as a core block for next-generation machine intelligent systems. 
	
The main contributions of this paper are:
\begin{itemize}
\item Develop a scalable and reconfigurable architecture for the HTM CLA.
\item Evaluate the performance of the spatial pooler and temporal memory for classification and prediction tasks.
\item Study the effect of noise on the classification and prediction accuracy of the HTM.	
\item Analysis of resource utilization for spatial pooler and temporal memory.
\end{itemize}	
	
The reminder of the paper is structured as follows: Section II presents an overview about the HTM theory. section III discusses the proposed hardware architecture. The simulation setup and verification methodology are discussed in section IV and V. Section VI demonstrates the results, while section VII concludes the paper.

\section{HTM Overview}
\begin{figure}
\begin{center}
\includegraphics[width = 0.3 \textwidth]{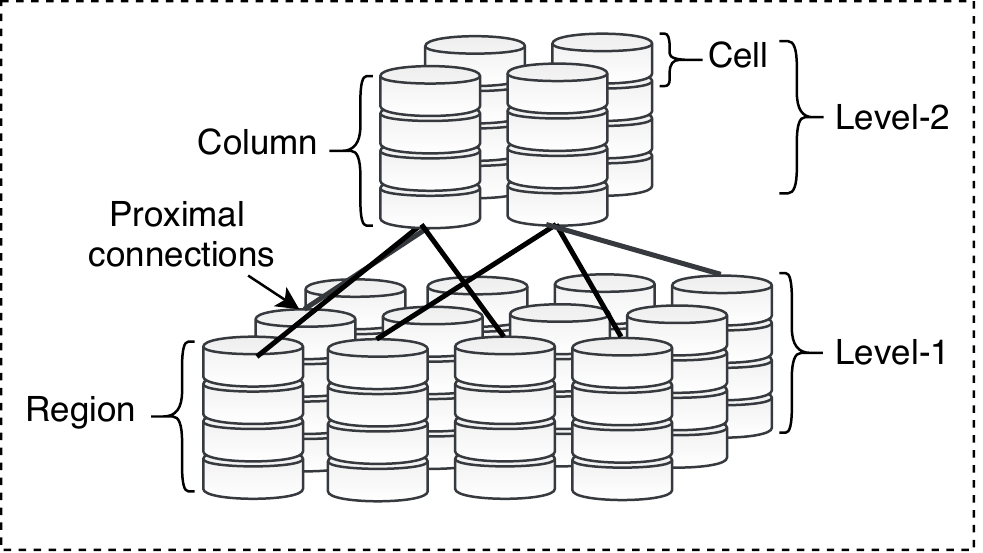}
\caption{A depiction of two HTM regions arranged hierarchically (two levels). Each region consists of columns of vertically stacked cells.}
\label{HTM_general_structure}
\end{center}
\end{figure}

The HTM network consists of regions or levels arranged in hierarchical form, illustrated in \fig{HTM_general_structure}, similar to the stratification in the neocortex. These regions are responsible for capturing spatial information and learning transitions between sequences. Each region within the hierarchy is composed of neurons known as cells, which incorporates the key capabilities of a biological pyramidal neuron. Within the same region, the cells are aligned vertically to form a columnar organization such that all cells within each column respond to only one particular feedforward input at a time. 

The cell represents the fundamental building block in HTM. Each cell has a \textit{proximal dendritic segment} and several \textit{distal dendritic segments}. The proximal segment connects the cells to either the input space or the layer immediately lower in the hierarchy, whereas the distal segments connect a cell to other neighboring cells within the same region. Each segment contains sets of synapses that are characterized by a scalar permanence value. The permanence value indicates the level of a synaptic connection strength and enable the system to learn and adopt. In the next subsections, the use of the cells and columns to realize the CLA are discussed.

\subsection{Spatial Pooler Model}
The SP is the first stage of the CLA. It is responsible for encoding the binary input data into a sparse distributed representation (SDR) using a combination of competitive Hebbian learning rules and homeostatic excitability control~\cite{cui2017htm}. The SDR is crucial to distinguish the common features between inputs~\cite{SDR_P}, learn sequences, and make predictions~\cite{sparse_whitepaper}. The SDR representation of the input is achieved by activating a small fraction of columns to represent the input patterns. Each column is connected to a unique subset of the input space using a proximal segment, which is composed of several synapses. For a given column, the cells share the same proximal connections. When a reasonable number of active synapses are connected to active bits in the input space, the proximal dendritic segment gets activated. The activation of the proximal dendritic segment will nominate that column to compete with its neighboring columns to represent the input. By using $k$-winner-take-all computation principle~\cite{lazzaro1989winner}, the column with the most active synapses and active inputs inhibits its neighbors and becomes active (winner). The output of the spatial pooler is as a binary vector, which represents the activation of columns in response to the current input. The length of the vector corresponds to the number of the columns and the ON bits refer to the winning columns. The operation of the spatial pooler can be divided into three distinct phases: initialization, overlap and inhibition, and learning.

\subsubsection{Initialization}
In this phase, which occurs only once, all the regions parameters are initialized including column's connections to the input space and synapses permanence. Let us assume the receptive field ($D_j$) defines a spot in the input space where the $j^{th}$ column synapses are connected; ($D_j$) is defined either locally or globally. In the local receptive field, the $j^{th}$ column synapses will be connected to a predefined region around its natural center in the input space, $\vec{x} \in \mathbb{R}^n_{\{0,1\}}$. In case of the global receptive field, the synapses of each column can be connected anywhere in the input space with no restrictions. The $j^{th}$ column connections, $\vec{c_j}\in \mathbb{R}^n_{\{0,1\}}$, to the input space can be found using~\eq{field}. 
\begin{equation}
\vec{c_j} = I(n_s, j \mid  j\mp \varepsilon ~\exists~D_j(r))
\label{field}
\end{equation}

where $I$ is an indicator function that returns a binary vector in which '1' indicates a potential synapse in the input space; $n_s$ and $\varepsilon$ are the number of potential synapses of the $j^{th}$ column and a random integer, respectively. $D_j$ is either a local receptive field centered at $j$ and has a radius $r$ or a global receptive field. The permanence of the $j^{th}$ column potential synapses ($\vec{\rho_j}$) is initialized with random values that are close to the permanence threshold ($P_{th}$), which is the minimum value of permanence that a potential synapse needs to be considered as a connected synapse. This speeds up the learning process because it activates the synapses after a few numbers of training iterations.

\subsubsection{Overlap and Inhibition}
The winning columns that represent the feedforward input are selected in this phase. This occurs after determining the overlap ($\alpha$) of each column within the input space. The overlap of the $j^{th}$ column is computed by counting its active synapses  that associate with active bits in the input space. Mathematically, it is achieved by performing a dot product operation, as in~\eq{overlap}:
\begin{equation}
\centering
\alpha_j = (\vec{c_j} \odot \vec{\rho_{j*}}) \cdot \vec{x}
\label{overlap}
\end{equation}
\begin{equation}
\centering
\vec{\rho_{j*}}= \text{I}(\vec{\rho_{j}} \geq P_{th})
\label{perm}
\end{equation}

where the active synapses vector is achieved through an element-wise multiplication (denoted as $\odot$) between the $j^{th}$ column potential synapses and the $j^{th}$ column permanence vectors. The permanence vector, $\vec{\rho_{j*}}$, is a binary vector to indicate the status of each potential synapse, where `1' refers to connected synapse and `0' unconnected synapse. The columns computed overlap ($\vec \alpha$) then get evaluated by comparing it to a threshold, known as minOverlap ($O_{th}$), as given by~\eq{eo}. The resulted vector is an indicator vector representing the nominated column with high overlap.
\begin{equation}
\centering
\vec{e\alpha} = \text{I}(\vec{\alpha} \geq O_{th})
\label{eo}
\end{equation}

Within a predefined inhibition radius ($\xi$), which can be local or global, the nominated columns compete against each other to represent the feedforward input. Based on the column overlap values and desired level of sparsity ($\eta$), $k_{th}$ number of columns will be selected to represent the input, as given by~\eq{kth}
\begin{equation}
\centering
\vec{w} = kmax(\vec{e\alpha}, \eta, \xi)
\label{kth}
\end{equation}

where kmax is a function that implements $k$-winner-take-all which returns the top $k_{th}$ elements within $\xi$.

\subsubsection{Learning}
After determining the winning columns, the learning phase starts to update the permanence values of the columns' synapses as necessary, but only the synapses of the active columns are updated. The approach followed in updating the permanence of the synapses is based on Hebbian rule~\cite{Hebb,hebb19880}. The rule implies that the connection of synapses to active bits must be strengthened, increase their permanence by $P^+$, while the connection of synapses to inactive bits must be weakened, decrease their permanence by $P^-$, as in~\eq{learn}.
\begin{equation}
\vec{\Delta\rho_{j}} = 
\begin{cases}
\vec{c_j} \odot \vec{\rho_{j*}} \odot \lambda \vec{x} - P^-, ~~ w_j = 1 \\
0 , ~~~~~~~~~~~~~~~~~~~~~~~Otherwise
\end{cases}
\label{learn}
\end{equation}

where $\vec{\Delta\rho_{j}}$ is the change in the permanence vector of the $j^{th}$ column and $\lambda$ denotes the sum of $P^+$ and $P^-$.
		
\subsection{Temporal Memory Model}
Learning sequences and making predictions for future inputs occurs in TM, the second stage of CLA. The cells of the winning columns are involved in this process. The active cells of the winning columns form lateral synaptic connections with the prior active cells. This allows the prediction of active state by just examining the distal segments. The number of distal segments that a cell might have depends on the number of distinct patterns that can be predicted. When a cell possesses higher number of distal segments, more connections with other cells can be formed and henceforth, more patterns can be predicted. The operation of the TM can be separated into three distinct phases: activate the cells of the columns, set the cells of the columns in the predictive state, and synapses permanence update.

\subsubsection{Activate the cells of the columns}
Each cell in the HTM is connected to other cells of the same region through its distal segments. The purpose of these segments is to determine the status of each cell at each point in time. Usually, the cells can be either in an inactive, predictive, or active state. In this phase, the cells are set to be in the active state. One cell per active column is selected to be active and learn the input pattern, if one of the column’s cells was in the predictive states in the previous time step. If the input is not predicted, i.e. there are no cells in the predictive state, all the cells of the active column are bursting to be in the active state. Furthermore, the cell with the largest number of active synapses, which is known as the best matching cell, is selected to be a learning cell to learn the input pattern. Let $a^t_{ij}$ denotes the activity of the $i^{th}$ cell in the $j^{th}$ column at time $t$, and $\pi^{t}_{ij}$ refers to the predictive state of that cell. The active state of the cell can be calculated as given by~\eq{segment}~\cite{hawkins2016neurons}.

\begin{equation}
a^t_{ij} = 
\begin{cases}
1,~\text{if}~ w^t(j)~\text{and}~\pi^{t-1}_{ij} = 1\\
1,~\text{if}~ w^t(j)~\text{and}\sum_i \pi^{t-1}_{ij} = 0\\
0 , ~~~~~~~~~~Otherwise
\end{cases}
\label{segment}
\end{equation}

\subsubsection{Set the cells of the columns in the predictive state}
The status of the distal segments of each cell is examined in this phase. The distal segments that are in the active state can turn its cell in the predictive state unless the cell is already activated by the feedforward input. The status of the distal segment can be determined after computing the number of active synapses connected to the active cell. Once that number exceeds the segment threshold ($S_{th}$), the segment will be in the active mode. When the distal segment turns into an active state, the segment synapses’ permanence are queued up to be updated in the next phase. The update of synapses occurs based on their cell status and whether they are connected to active or inactive cells in the same region. Let A denotes the activity of cells in the region, $DS^d_{ij}$ is a $d^{th}$ distal segment of $i^{th}$ cell within the $j^{th}$ column, where A and $DS^d_{ij}$ are binary matrices of size $n_c$x$n_m$ ($n_c$ and $n_m$ denote the total number of columns and cells in a region, respectively). Thus, the predictive state of a cell at the current time step is given by~\eq{predict}~\cite{hawkins2016neurons}.

\begin{equation}
\pi^{t}_{ij} = 
\begin{cases}
1, ~\text{if}~\exists_d~\lVert DS^d_{ij} \cdot A^t \rVert_1 > S_{th} \\
0 , ~~~~~Otherwise
\end{cases}
\label{predict}
\end{equation}

\subsubsection{Synapses permanence update}
The update of the segments that are queued up in the second phase of TM is carried out in this phase (learning phase). The cells that are in the learning state and correctly predicted, its distal segments are positively reinforced, whereas the cells that stop predicting its segments are negatively reinforced.

\begin{figure*}
\centering
\includegraphics[width=0.7\textwidth]{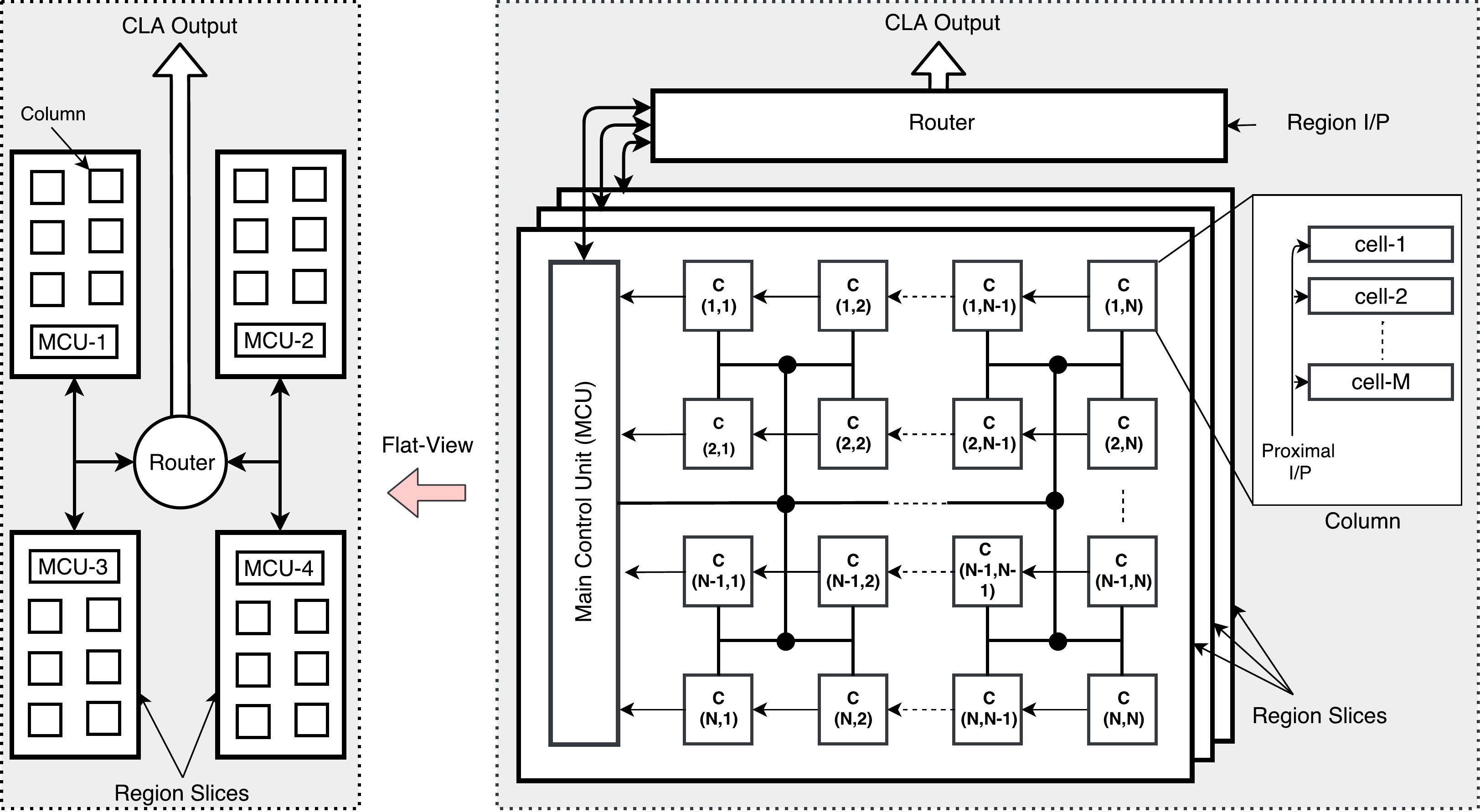}
\caption{Block diagram representation of HTM network architecture. The architecture is composed of columns of cells that emulate the biological pyramidal neuron, a MCU to enable columns within a region slice to communicate with others and to generate the region slice final output, and a router to bridge the region slices together and with other levels in HTM hierarchy including the input space.}
\label{HTM_pipelined}
\end{figure*}

\section{Proposed HTM Architecture}
\fig{HTM_pipelined} illustrates the high-level architecture of the proposed HTM region, comprising of region slices and a router. The region slices are used to develop a scalable size of HTM regions\footnote{In this work, one region slice is considered as one HTM region and the router is used just to demonstrate the concept of scalability. Hence, the router will not considered in the discussion and the MCU will be the unit that bridges the HTM region to the encoder.}, whereas the router enables the interaction between the region slices and bridges them to input space (encoder) and other regions in the hierarchy. The region slice (or region in short) consists of several homogeneous columns, each with vertically aligned cells. There is an inherently parallel nature in the column operation. To exploit this innate structure, the proposed design is pipelined and has parallelism. Such reconfigurability and flexibility in the design eases portability onto different FPGA fabrics. At a high-level, an HTM region consists of three modular building blocks, the column, the cell, and the main control unit (MCU), discussed in the following subsections.

\begin{itemize}
\item \textbf{Column:} A pipelined 2-D array represents the HTM columns as shown in~\fig{HTM_pipelined}. Each column in the region is responsible for computing its overlap value and updates its synapses' permanences during the learning phase. Each column receives input data from the MCU and computes its overlap score locally. The overlap information is sent to the MCU in a pipelined fashion to determine the winning columns. This is followed by the update of column synapses' permanences which occurs locally in columns as well. 

\item \textbf{Cell:} The cells are embedded within the columns as minor cores that receive data from columns. Based on the history and the current input, a cell establishes lateral synaptic connections with other cells in the same region to learn sequences and thereby predictions. This happens after determining the winning columns and starting the second phase of the CLA algorithm, TM. 

\item \textbf{MCU:} The MCU functions as a bridge in order to enable communication between HTM regions and the encoder. It also manages the data flow between pipeline stages, thereby maintaining sequence. This reduces data bus width to the MCU as there is only transfer of overlap values (without columns' numbers). After receiving all the overlap scores, the MCU selects the winners using a $k$-winner-take-all circuit. In the $k$-winner-take-all circuit, the top 20\% columns are selected to compose the input SDR. The selection is based on the level of activation (overlap value). A column has a higher chance of winning if it has a higher activation value. During the TM phase, the MCU is responsible for dispersing the network information so that each cell knows the current active cells. This process enables the inactive cells that have active distal segments making predictions. At the end of this process, the MCU generated the final output of the HTM region as a binary vector that holds the status of each cell.
\end{itemize}

\begin{figure} [h!tb]
\centering
\includegraphics[width = 0.4\textwidth]{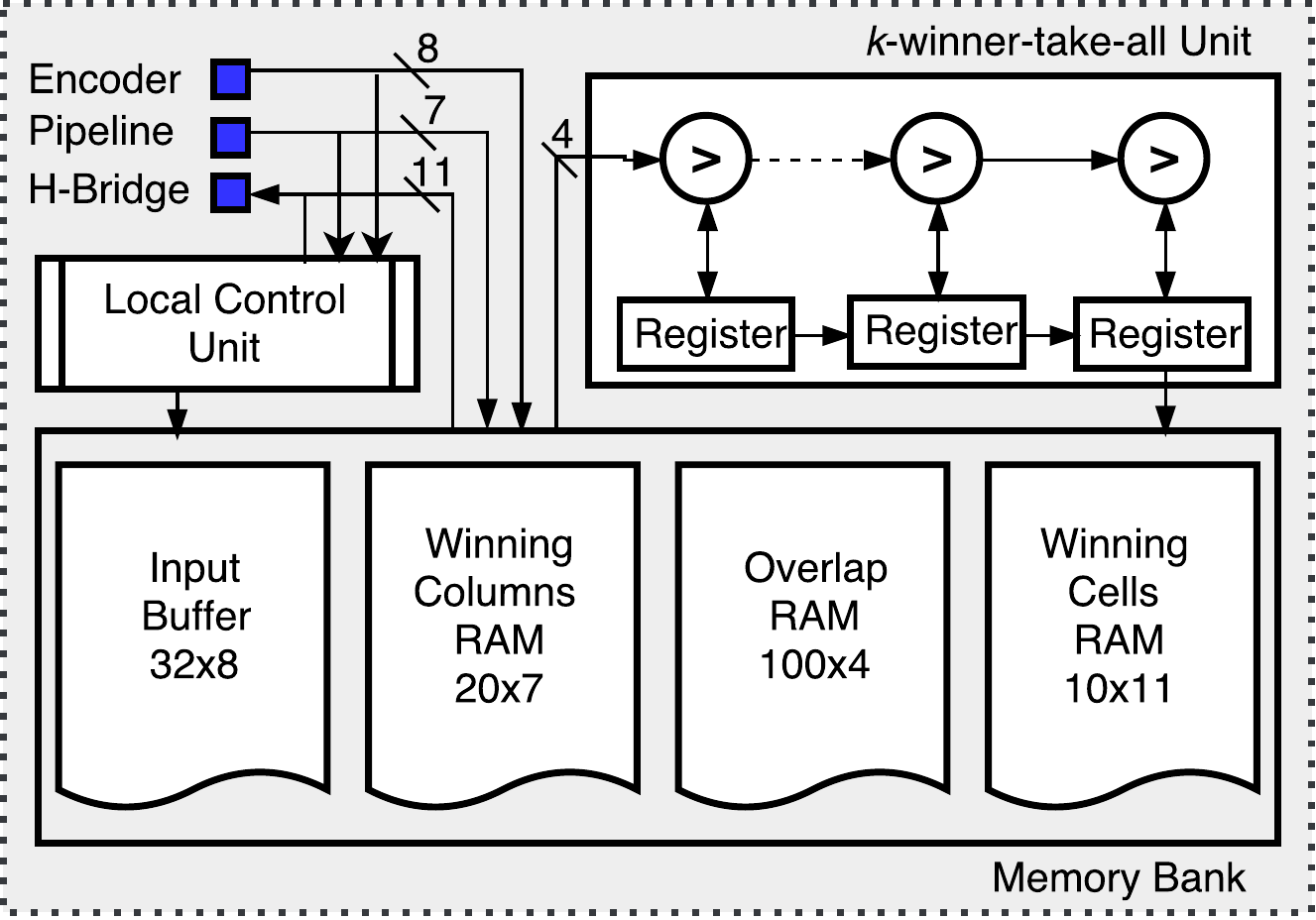}
\caption{MCU RTL diagram. It is composed of a memory bank to store columns and cells information, $k$-winner-take-all circuit to select the winning columns, and a local control unit.}
\label{fig:HTM_control_unit}
\end{figure}
			
\subsection{MCU RTL Design}
The MCU bridges the HTM columns to the input encoder or other regions in the hierarchy. Moreover, it is responsible for dispersing data, selecting the winning columns, and generating the SP and TM final outputs. In this section, we focus on the compatibility between the MCU and the columns array. The MCU receives data from the encoder (\fig{fig:HTM_control_unit}) sequentially in the form of binary packets, where a packet refers to a single byte in the input vector. Once a packet is received, it is forwarded to the memory bank and stored in the input buffer. The input packets will be relayed to the columns via the H-Bridge network so that each column computes its overlap value. The computed overlap values are sent back to the MCU via the pipeline buses and stored in the \textit{Overlap RAM} within the memory bank. Then, the overlap values are randomly extracted from the \textit{Overlap RAM} to be fetched in the $k$-winner-take-all unit. The $k$-winner-take-all unit consists of a series of comparators and registers which are used to pick the top 20\% column overlaps and store them in the registers. The output of the registers is in turn passed to the \textit{Winning Col.} RAM. Then, the list of the winners will be sent back to the columns to update their synapse permanences during the learning phase. After the learning phase of the SP, the TM starts, in which the MCU plays the role of information distributor. Typically, the TM starts by selecting the winning cells that represent the input within their context. Once the cells are determined, they will be sent to the MCU via the pipelined network formed by the columns. The MCU is responsible for receiving this data and dispersing it back to the network so that each cell knows the current active cells. This process enables the cells, which are in inactive mode and have active distal segments, making predictions. While the cells are busy, the MCU makes use of this time and generates the final output of the HTM region and starts the next iteration.

\subsection{Column RTL Design}
The cells in the column are activated due to feedforward input or the lateral input of the nearby cells. Designing such dynamical connections in hardware is challenging because of the rigid interconnections and thereby lack of the flexibility. To overcome this limitation, a synthetic synapse concept is developed in this work. Synthetic synapses describe the characteristics of the physical interconnects such as its address and permanence using a memory unit (or LFSR + Memory) associated with each column. Although this realization  increases the computation time, it significantly reduces the interconnect area and offers dynamic connectivity to port onto the FPGA.

\begin{figure} [h!tb]
\centering
\includegraphics[width = 0.4\textwidth]{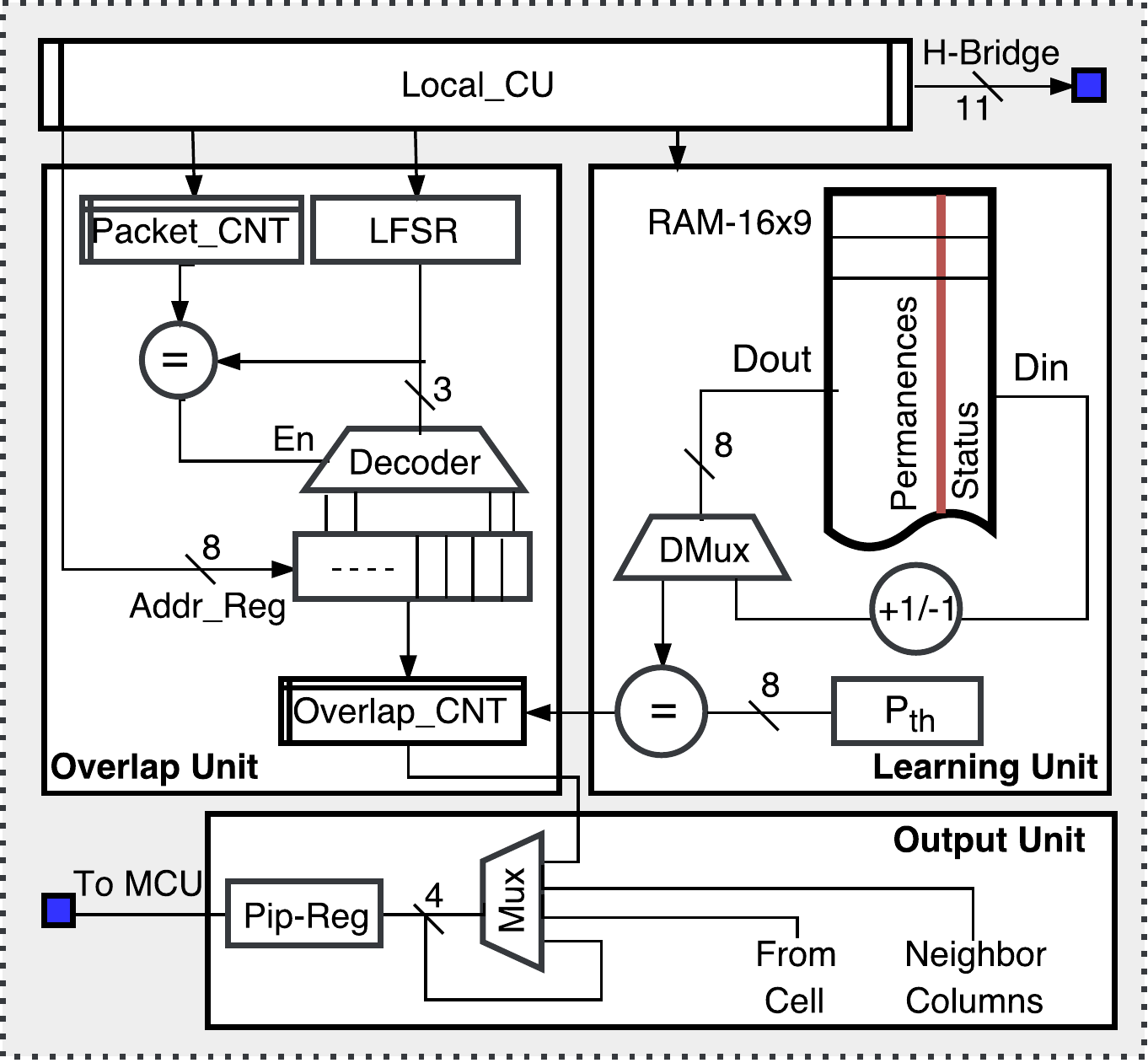}
\caption{RTL representation of the HTM-column (without cells), which is composed of four main units. Overlap unit to compute the overlap value, a learning unit to store the synapse permanence value and to perform the learning process, output stage to connect the columns and its associated cells to the pipeline network, and a local control unit.}
\label{fig:HTM_column}
\end{figure}   

During the initialization phase, all columns memory units will be loaded with the synapse permanence values. Then, the synapses' addresses in the input space are located. Typically, the synapses' addresses are either generated by using an LFSR or stored in a ROM, which has previously generated receptive fields. In the LFSR version, each column can establish synaptic connections anywhere in the input space without restrictions. This approach gives all the columns the same likelihood to be active. Moreover, the column activation over time will be more consequent so that all the columns take part in representing the feedforward input and avoid any dead columns. The second approach, which is called ROM receptive field (RRF), is captured from~\cite{cla_whitepaper} in which each column has a receptive field around its center in the input space and the neighboring columns are sharing some locations in the input space. Although this method can give better encoding results in terms of giving similar input similar representations, the columns that are connected to the low active areas in the input space are less likely to be active. For this reason, these column activations need to be \textit{boosted} up often to participate in representing the input data.

Upon completion of the initialization phase, the columns are then ready for the \textit{overlap phase}. This phase starts by loading the \textit{Addr\_Reg} register with an input packet sent from the MCU via the H-Bridge network (shown in~\fig{fig:HTM_column}). Then, a synapse address will be generated (or loaded in case of RRF) and decoded to observe for input matching. If the synapse address in the input space corresponds to '1', its permanence is observed. Whenever the synapse's permanence is more than the threshold ($P_{th}$) and the synapse address in the input vector is logic '1', the overlap counter (\textit{Overlap\_CNT}) will be increased by one. Otherwise, a new synapse address is generated. This process continues until all the synapse addresses are examined with their corresponding input bit addresses. It is important to mention here that during the overlap phase, the synapses' status (active or inactive) is registered in the permanence RAM. This step will facilitate the learning process in the next phase. Once all the columns' overlaps are computed, they will be transferred via the pipeline register to the MCU in order to select the winning columns. While the MCU determining the winning columns, all columns will be in a steady state. Then, the MCU, through the H-Bridge, broadcasts the list of the winning columns to the region. Each column checks if it is a winner. If so, the column updates its synapses permanence according to to the Hebbian rules~\cite{Hebb}. If the synapse status, as registered during the overlap phase, is '1', the connection is strengthened via \textit{($+1/-1$)} unit, otherwise it gets weakened. For the rest of the columns, i.e. not winners, their synapses' permanence will remain unchanged.  


\subsection{Cell RTL Design}
The column cells are responsible for representing the input data within the context and making predictions. This occurs during the TM phase of the CLA algorithm. Typically, to develop an HTM network that can predict high order sequences multiple cells are required. At every point in time, one of these cells is selected to be active. This feature can be exploited to simplify HTM hardware architecture. Rather than realizing multiple individual cells per column, a single cell with partitioned memory can be used.

The operation of the cell unit begins during the TM. It starts with the initialization process in which the synapses’ permanences are initialized around the permanence threshold. To simplify the hardware, it is suggested to initialize all the synapses with the same permanence value, $P_{th}$-2, which is loaded into \textit{Permanence} partition. After the initialization, all cells are set to be in the ideal state until the winning columns of a particular input are determined. Each column informs its cells about its status via the command bus, which is 3-bit wide. The cells decode the received command via a set of AND gates and perform the necessary tasks. If the column is a winner, it is expected to receive ``111" through the command line, otherwise, ``101" is received.

When the cell unit receives a flag from its column, phase-1 starts by examining the cells’ status in the previous time step, which is stored in \textit{CellsTimeLine} memory. For a given column, if one of its cells was in the predictive state, this cell is selected to be active (in case the cell is set to be active due to active cells that were in the learning state, this cell is selected to be a learning cell as well). When none of the column's cells are in the predictive state, bursting will happen and all the cells of the column will be ON. Then, the best matching cell, the one with the largest number of active synapses, is chosen to learn the current feedforward input. This occurs in the \textit{Burst Block}, where the best matching cell will be stored in \textit{ch\_learn}. The number of active cells will be transferred to the column and then to the MCU. The MCU re-distributes the winning numbers to all the cells in the network. If a cell is selected as a winning cell, it starts phase-3 to establish a new segment or updates its synapse permanence for learning, otherwise phase-2 starts.

\begin{figure}[ht]
\centering
\includegraphics[width = 0.45 \textwidth]{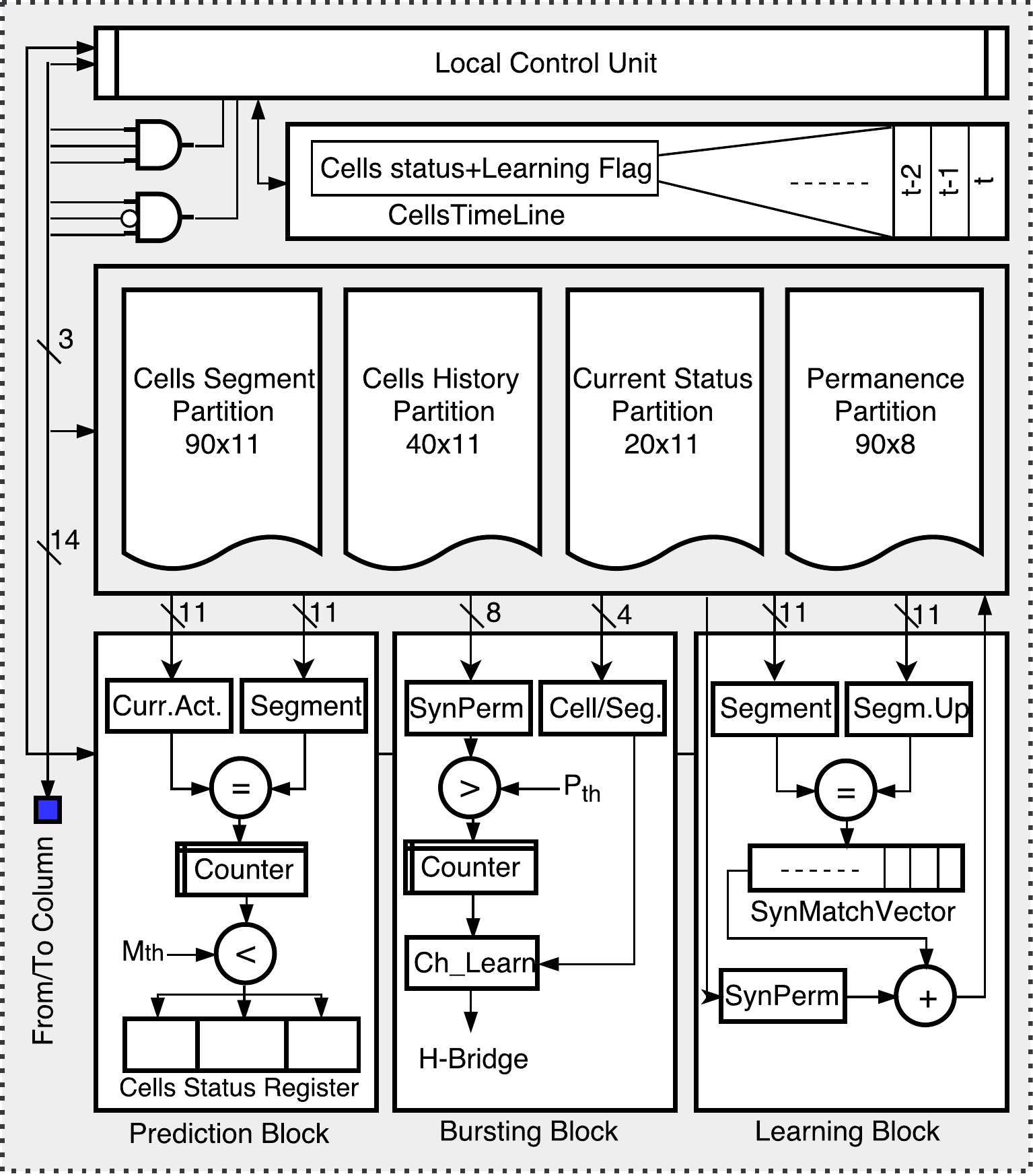}
\caption{RTL representation of multiple cells (chosen to be 3 in this work) within a column of the HTM, which is mainly composed of a memory bank to store cells data, 3 computational blocks to execute the TM phases, and a local control unit.}
\label{fig: cell_HW}
\end{figure}

Phase-2 is responsible for setting the network cells in the predictive state and updating the \textit{CellsHistory} partition to add the current active cell locations and the ones prior (stored into \textit{Current Status}). Whenever a cell is not in an active state and has one or more active segments, the prediction occurs. This happens in \textit{Prediction Block} in \fig{fig: cell_HW}. This process starts by reading the stored segment addresses and the current active cell addresses and checks for matching percentage ($M_{th}$). Whenever there is a reasonable match (it is selected to 50\% in this work), the cell of the matched segment is set to be in the predictive state by setting its status to ``11". Otherwise, none of the cells will be in the predictive state and the current active segment will be adopted in the next time step in phase-3. Once phase-2 or phase-1 is finished, phase-3 will start. This phase starts by checking the cells’ status in the previous and current time step. If one of the cells is in the predictive state and currently active, this means the current input matched the predicted and the synapses permanence will be strengthened. Yet, if there was a cell in the predictive state and currently inactive, its synapses permanence will be weakened. However, in the \textit{Learning Block}, the \textit{CellsSegment} partition is compared with the \textit{CellsHistory}. When there is a match, the \textit{SynpMatchVec} will be filled with the location of the matched addresses to strengthen their connection, otherwise, it will be transferred to the Segment partition.

\section{HTM Simulation Setup and Verification Datasets}
\subsection{HTM Network Parameters Setup}
The performance of HTM network, its structure and utilized hardware resources (see Table \ref{SPParameter}), are highly affected by the network parameters. This section will discuss the techniques adopted to set various HTM network parameters.

\begin{table}[ht]
\caption{The spatial and temporal memory parameters used for simulations.}
\label{SPParameter}
\begin{center}
\begin{tabular}{ | l | c |  }
  \hline                   
  \rowcolor{Gray} \textbf{SP parameters} & \textbf{Value} \\ \hline
  Number of columns ($n_c$) & 100 \\ \hline
  Sparsity level ($\eta$) & 20\% \\ \hline
  Number of synapses per column  & 16 \\ \hline
  Synapse permanence threshold ($P_{th}$)& 127 \\ \hline
  Permanence dec/inc factor ($P^-/P^+$) & 1 \\ \hline
  Inhibition type & Global \\ \hline
  MinOverlap ($O_{th}$)& 2 \\ \hline \hline 
  \rowcolor{Gray} \textbf{TM parameters} & \textbf{Value} \\ \hline
  Number of cells per column ($n_m$) & 3 \\ \hline
  Number of segments per cell & 3 \\ \hline
  Number of synapses per segment & 10 \\ \hline
  Permanence dec/inc factor ($P^-/P^+$) & 1 \\ \hline
\end{tabular}
\end{center}
\end{table}

\subsubsection{Synapses' parameters} According to~\cite{cla_whitepaper}, 
the permanence of the synapses is randomly initialized to values near the permanence threshold in order to establish active connections after a small number of training epochs. The permanence is a real-scalar value bounded between [0,1]. To mitigate usage of complex and resources-intensive floating or fixed point units in the hardware, the permanence value is set between [0-255], with a midpoint threshold of 127. 
	
\subsubsection{Overlap threshold} The minOverlap parameter, selected experimentally, controls whether or not a column is considered in the inhibition process. It is recommended to keep it within a range so that enough number of columns can be activated when the input is presented to the network, and at the same time the columns are not activated by trivial patterns or noise.
		
\subsubsection{Number of columns} In order to meet the network minimum constraints, such as maintaining the sparsity level and the semantic meaning, the number of columns selected for the test platform is 100 ($n_c$), while the number of active columns is no more than 20 ($w_c$) at any given point of time.  This can give a sparsity level ($\eta$) of 20\% as in (\ref{sparsity_level}). 
\begin{equation}\label{sparsity_level}
\eta  = \dfrac {w_c}{n_c} \times 100
\end{equation} 
 
For the given network setup, the network can generate up to $5.36 \times 10^{20}$ unique representations as given in \eq{SDR encoding}~\cite{sparse_whitepaper}. While the sparse encoding capacity ($\psi$) is clearly less than that seen in dense encoding, this difference tends to be meaningless as the number of columns grow~\cite{sparse_whitepaper}. 
\begin{equation}\label{SDR encoding}
{\psi  = \frac{n_c!}{w_c!(n_c-w_c)!}} \\
\end{equation}
	
\subsubsection{Number of synapses} A total of 16 synapses are utilized per column to ensure each column has a reasonable overlap with the feedforward input such that generating trivial patterns is avoided.

\subsubsection{Number of cells} The number of cells is chosen to be 3 per column. This enables the network to represent more sequences and make more predictions.

\subsection{Verification Datasets}
In order to validate the proposed HTM-SP architecture, the standard hand-written digits, MNIST, is used. The dataset consists of 60,000 training examples and 10,000 testing examples where each example is a gray-scale image of 28x28 pixels. These images are binarized and resized to 16x16 pixels before being utilized in the HTM network. This is to ensure that the training time and amount of data stored in the hardware will be reduced. The second dataset, which is used to evaluate the performance of the HTM-TM is the numbers of the European numberplate font (EUNF)~\cite{EU_Numbers}. The numbers of the set,  which  range  between  0-9, are resized to 19x14 pixels and binarized so that it can be used with HTM network.

\section{Verification Methodology}
\subsection{Spatial Pooler Verification}
The main fundamental task of the SP is to encode the input data into an SDR with fixed sparsity. The encoding process is subjected to constraints such that similar inputs are mapped to similar or at least close representations. In order to verify whether the SP performs the intended functions, two verification steps are performed. The first step checks if the SP generates sparse representations. This test is easy and can be applied by utilizing any traditional debugging method. The other set focuses on observing the semantic meaning within the generated SDRs. This part is performed using SVM classifier and MNIST dataset. It starts by setting the SP parameters and pre-processing the MNIST images by the encoder. Then, the output of the encoder is presented to the network for training, testing, and observing the classification accuracy (SVM classifier output). Different sets of HTM parameters are observed and its influence on the classification accuracy will be discussed in the results and analysis section.

\subsection{Temporal Memory Verification}
The TM is developed to learn sequences and make predictions, consequently, it is verified with a prediction task. Unlike the SP, the TM is too difficult to debug traditionally. Thus, a software/hardware debugging environment is developed. This reads the cells' memory, where all the significant information associated with columns' cells are stored, and save them in a CSV file. Then, the cells' information stored in the CSV file are mapped to a 3D plot, which can be used to visualize an HTM region graphically and to facilitate the debugging process. The 3D plot has 3 layers of cells aligned vertically and arranged in a 2D array. The software code reads the status of the cells at the current time step and the two before, then maps it to the 3D plot so that the status of the cell in each time step can be visually seen and easily tracked. 

\begin{figure} [h!tb]
\begin{center}
\includegraphics[width = 0.4 \textwidth]{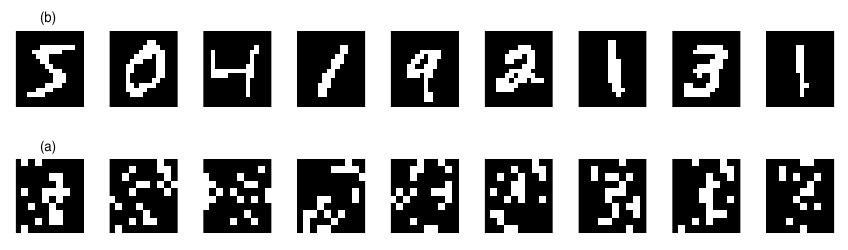}
\caption{Visualization of MNIST images and their corresponding SDRs as generated by the SP.}
\label{MNIST_Graph_2}
\end{center}
\end{figure}
\section{Results and Analysis}

\begin{figure*}[h!tb]
\centering
\subfigure[]{\includegraphics[width=50mm, height=40mm]{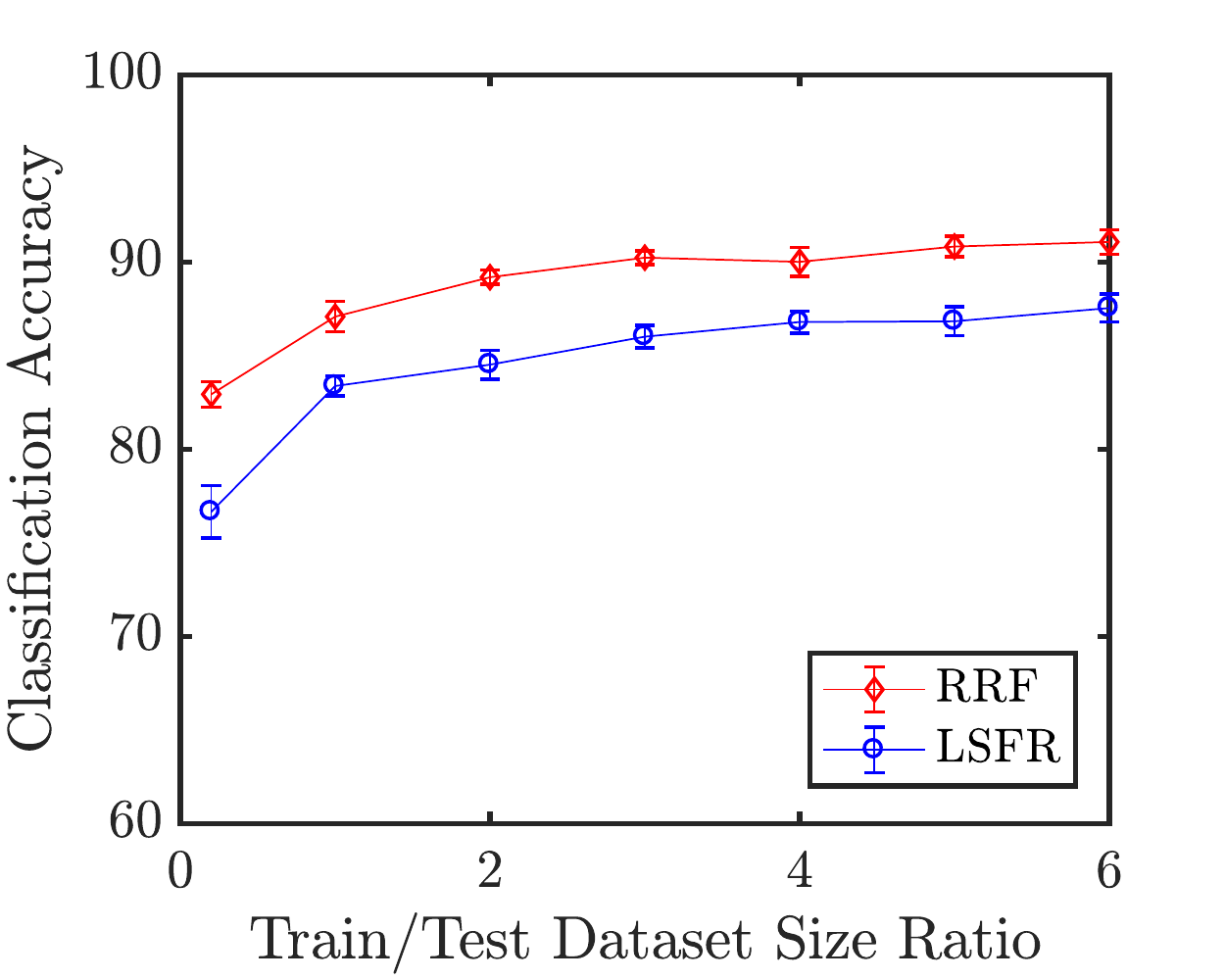}}
\subfigure[]{\includegraphics[width=50mm, height=40mm]{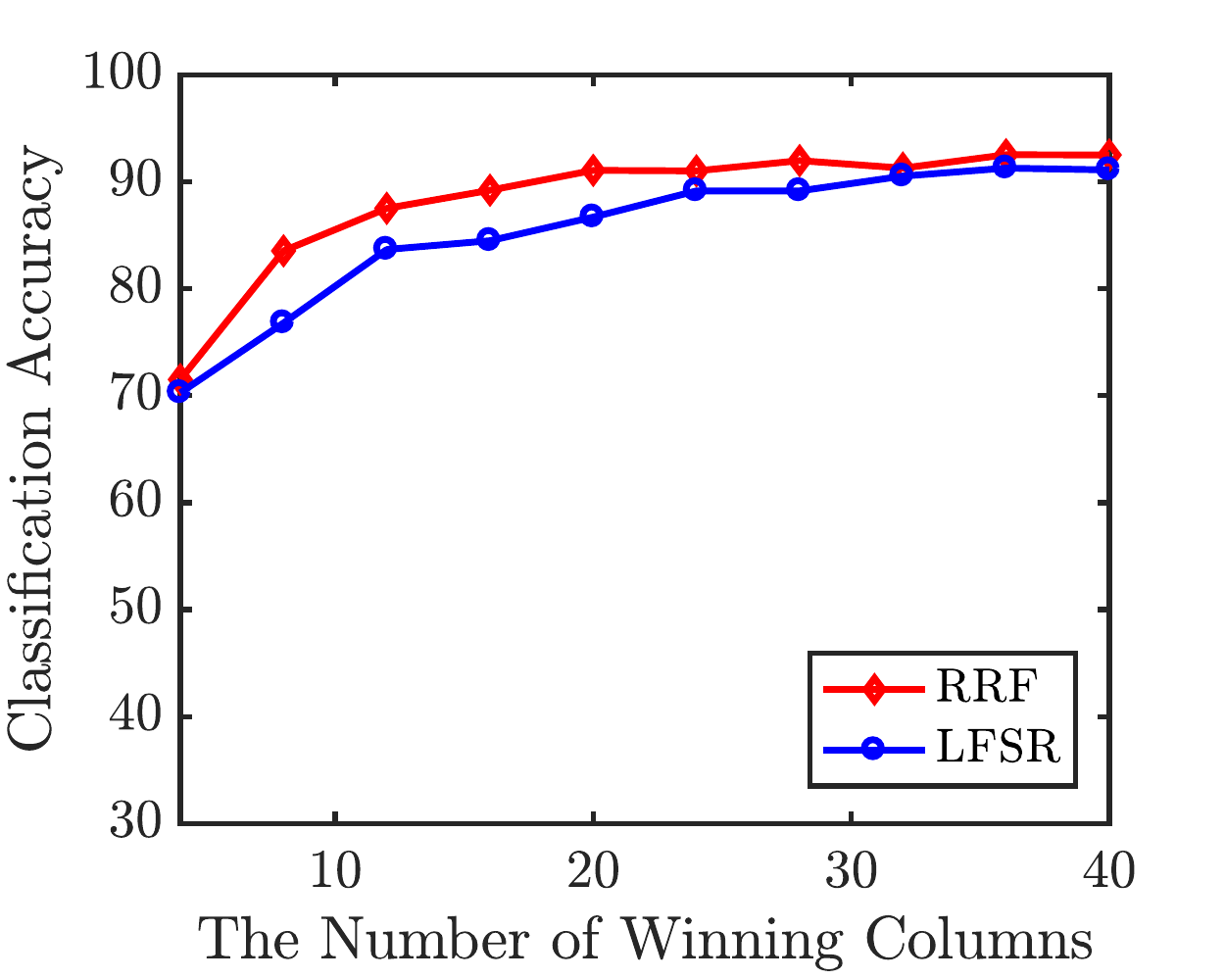}}
\subfigure[]{\includegraphics[width=50mm, height=40mm]{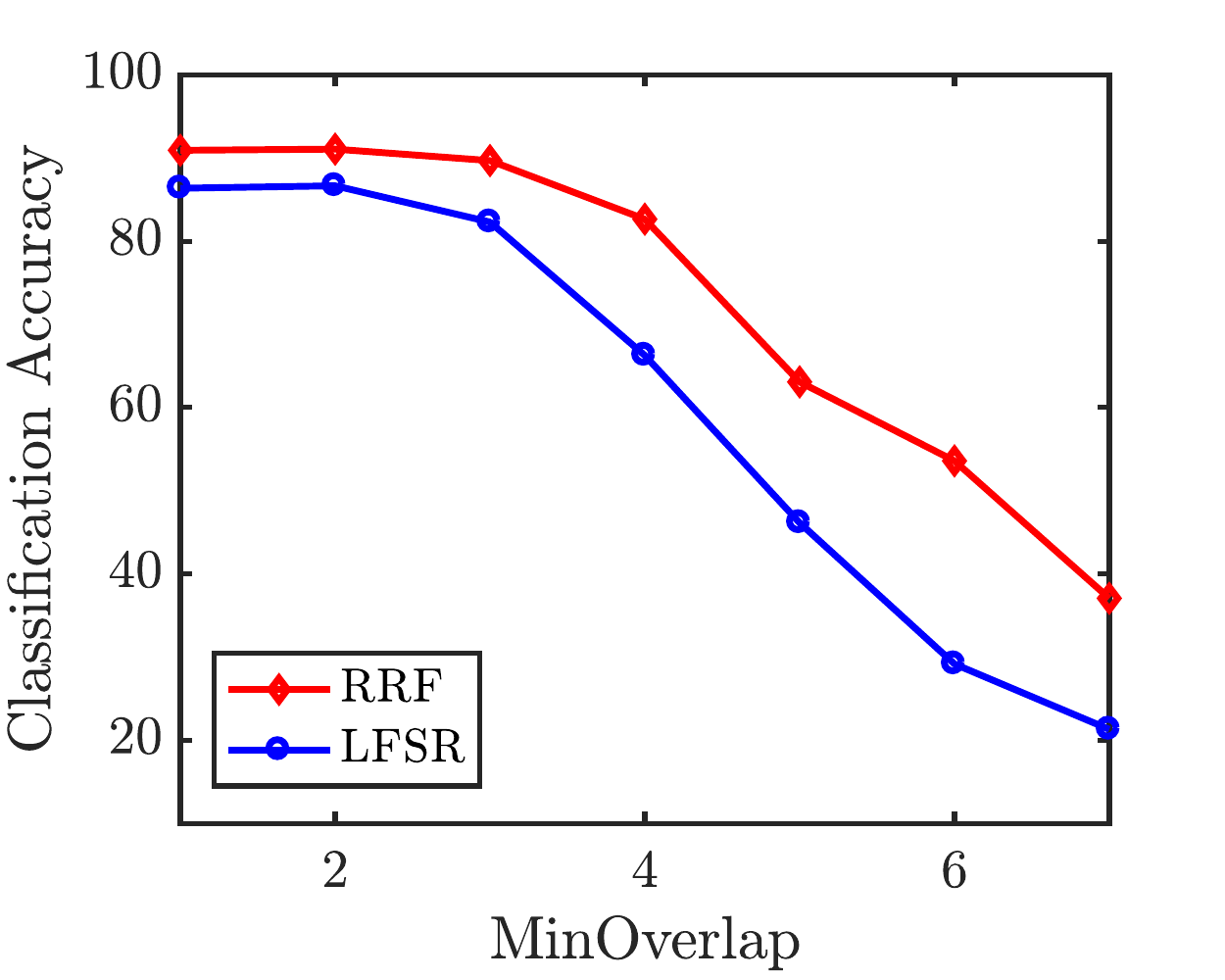}}
\caption{(a) The impact of training set size on SP+SVM classification accuracy. (b) SP+SVM classification accuracy of LFSR and RRF (receptive fields) for a various number of winning columns. (c) The effect of varying the overlap threshold (minOverlap) of SP+SVM Classification  accuracy.}
\label{accur}
\end{figure*}

\subsection{Classification Accuracy} 
In order to validate the proposed HTM-SP architecture, MNIST and EUNF datasets are used. First, the SP is trained and tested in unsupervised manner on these datasets separately, then its output (shown in \fig{MNIST_Graph_2}) is classified with a supervised classifier (SVM and k-nearest neighbor (KNN) are used in this work).~\tb{accuracyTable} depicts the classification accuracy achieved and compare it with~\cite{james2017htm,streat2016non,woods2015synaptic}. When using SP+SVM classifier, this work offers comparable accuracy to~\cite{james2017htm,streat2016non} with less number of columns. This may attribute the use of deterministic receptive fields which enables HTM region maintaining the semantic meaning of the feedforward input even with a small number of columns. When comparing SP+SVM with other sparsity classifiers, such as LCA in~\cite{woods2015synaptic}, it is found the SP+SVM outperform the LCA by $\approx$1\%. \fig{accur}-(a) shows the test classification accuracy with standard deviation error bar of MNIST dataset as we changed the training set size and kept the test set size fixed. It can be noticed that the accuracy of the network is gracefully degraded even when the ratio of training/testing dataset is 0.2 (first iteration). This indicates the capability of the HTM-SP to learn from a small set of examples and generalize it for the rest. This is added to the stable performance even when the receptive field of the columns are changed. More detailed analysis of the classification accuracy with varying parameters is presented in the following subsections.

\begin{table}[ht]
\caption{Summary of classification accuracy of MNIST dataset.}
\label{accuracyTable}
\centering
\begin{threeparttable}
\begin{tabular}{|c|c|c|c|}
\hline
\rowcolor{Gray} \textbf{Work} & \textbf{No. of columns} &\textbf{Classifier} & \textbf{Accuracy} \\ \hline 
mHTM~\cite{mnatzaganian2017mathematical}\tnotex{tnote:robots-r1} & 936 & SP+SVM  & 92.39\% \\  \hline
F-HTM~\cite{streat2016non} & 784 & SP-SVM  & 91.98\% \\ \hline 
Memristive-LCA \cite{woods2015synaptic} & 300 & LCA  & 90.0\%  \\ \hline
This work & 100 & SP+SVM  & 91.16\%\\ \hline
This work & 100 & SP+KNN  & 87.47\%\\ \hline
\end{tabular}
\begin{tablenotes}
\item\label{tnote:robots-r1} Software implementation of HTM-SP.
\end{tablenotes}
\end{threeparttable}
\end{table}

\subsubsection{HTM Parameters Vs. Classification Accuracy}
In order to investigate the influence of the parameters tuning on how the SP may encode the feedforward input data, several parameters are tuned to observe this effect. These parameters are the number of winning columns and the \textit{MinOverlap}. \fig{accur}-(b) demonstrates the impact of the number of winning columns on the classification accuracy. It can be seen that as the number of the winning columns increases, the classification accuracy gets improved and this attributes to the encoding capacity improvement resulted from increasing the number of columns. \fig{accur}-(c) shows how the classification accuracy is inversely proportional to the \textit{MinOverlap}. The \textit{MinOverlap} has a significant influence on the number of active columns that participate in the input representation. With higher \textit{MinOverlap}, fewer columns will be able to pass the \textit{MinOverlap} and a fixed level of sparsity will be hard to achieve. This eventually leads to degrade classification accuracy.

\subsubsection{Noise effect on Classification Accuracy}   
This section evaluates the robustness of the SP performance in encoding the feedforward input with the presence of noise. Two datasets are utilized for this purpose: EUNF and MNIST. For both datasets, the images are resized as aforementioned after injecting the noise. Then, all the images are binarized so that it can be processed by HTM region. In this work, Gaussian and Salt\&Paper noises are used to distort the dataset images. \tb{NoiseTable} shows the noise type, its intensity, and its effectiveness on the classification accuracy. It can be noticed that the noise level has a noticeable impact of the classification accuracy which attributes to the results of false representation that may be generated due to the injected noise. However, the results show that the system can handle noise with a density of 20\%.

\begin{table}[ht]
\caption{Impact of various types of noise on the SP+SVM classification accuracy for EUNF and MNIST dataset.}
\label{NoiseTable}
\begin{center}
\begin{tabular}{|c|c|c|c|}
  \hline              
 \rowcolor{Gray} \textbf{Noise Type} & \textbf{Noise Den. or Var.} & \textbf{Acc. EUNF} & \textbf{Acc. MNIST} \\ \hline
  Salt\&Pepper & 10\% density & 100\% & 86.25\% \\ \hline
   Salt\&Pepper & 20\% density& 90\% & 84.38\% \\ \hline
   Gaussian & 10\% variance & 100\% & 86.45\%\\ \hline
   Gaussian & 20\% variance & 90\% & 83.70\%\\ \hline
\end{tabular}
\end{center}
\end{table}

\begin{figure*} [h!tb]
\centering
\subfigure[]{\includegraphics[width = 0.2 \textwidth]{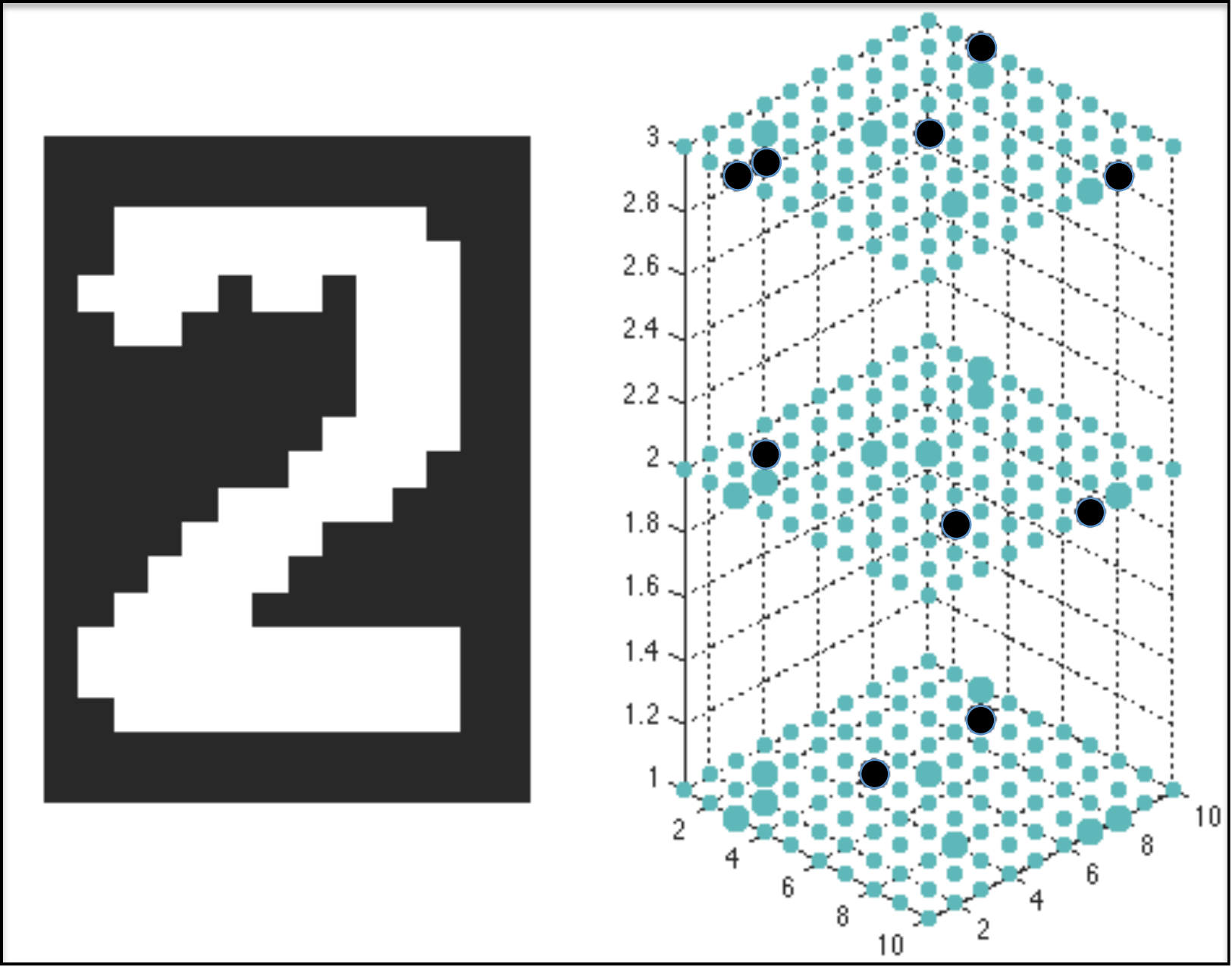}}
\subfigure[]{\includegraphics[width = 0.2 \textwidth]{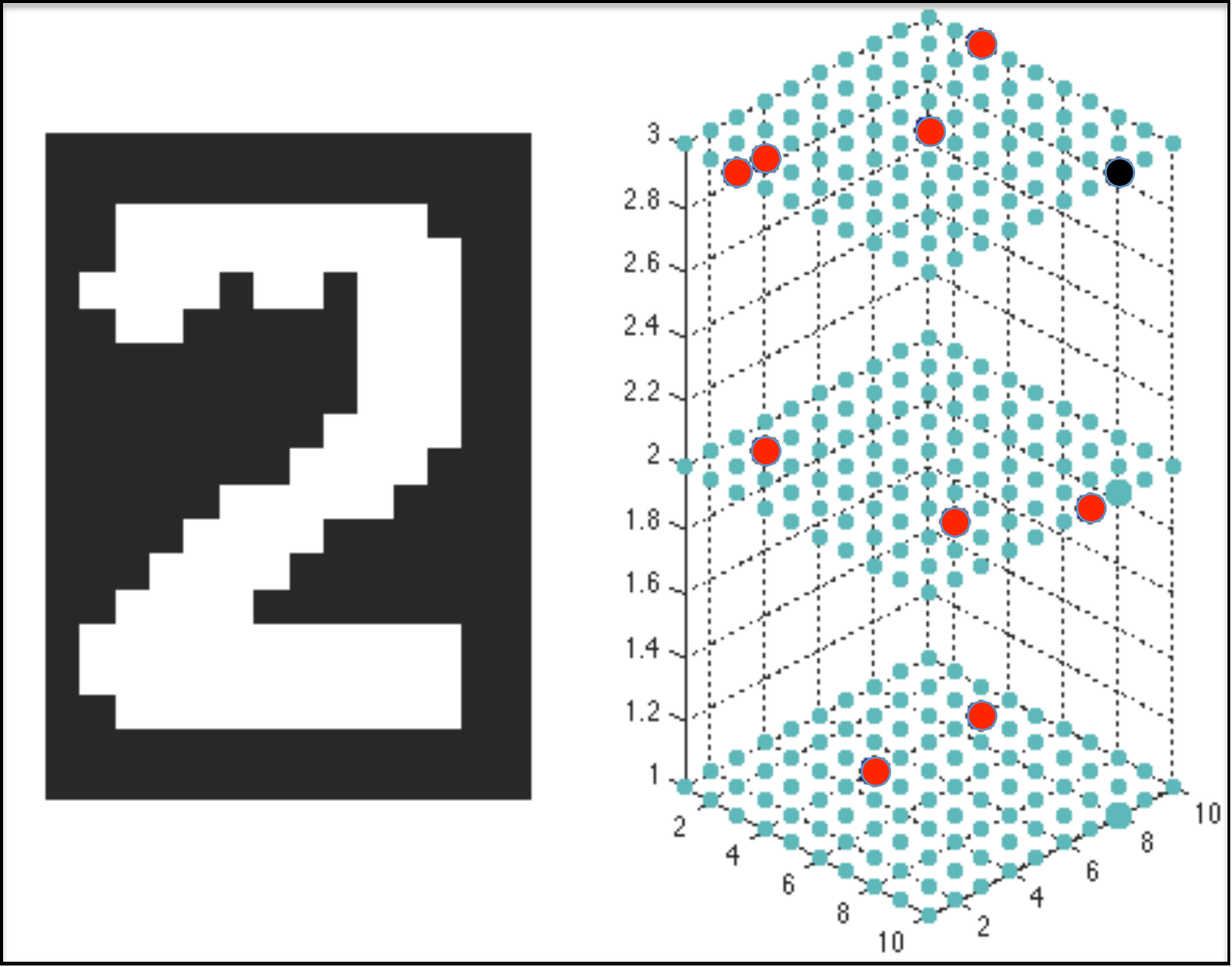}}
\subfigure[]{\includegraphics[width = 0.2 \textwidth]{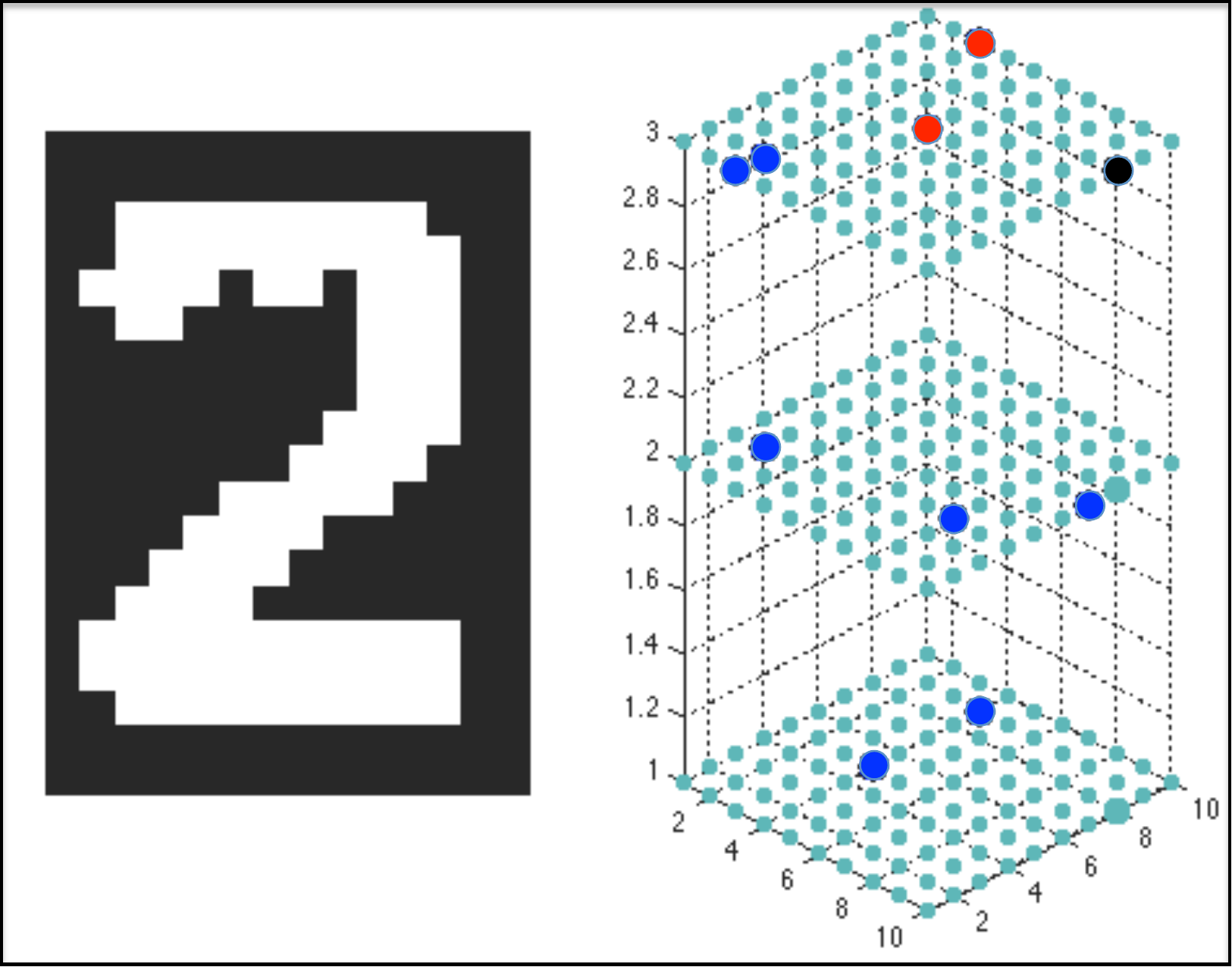}}
\subfigure[]{\includegraphics[width = 0.2 \textwidth]{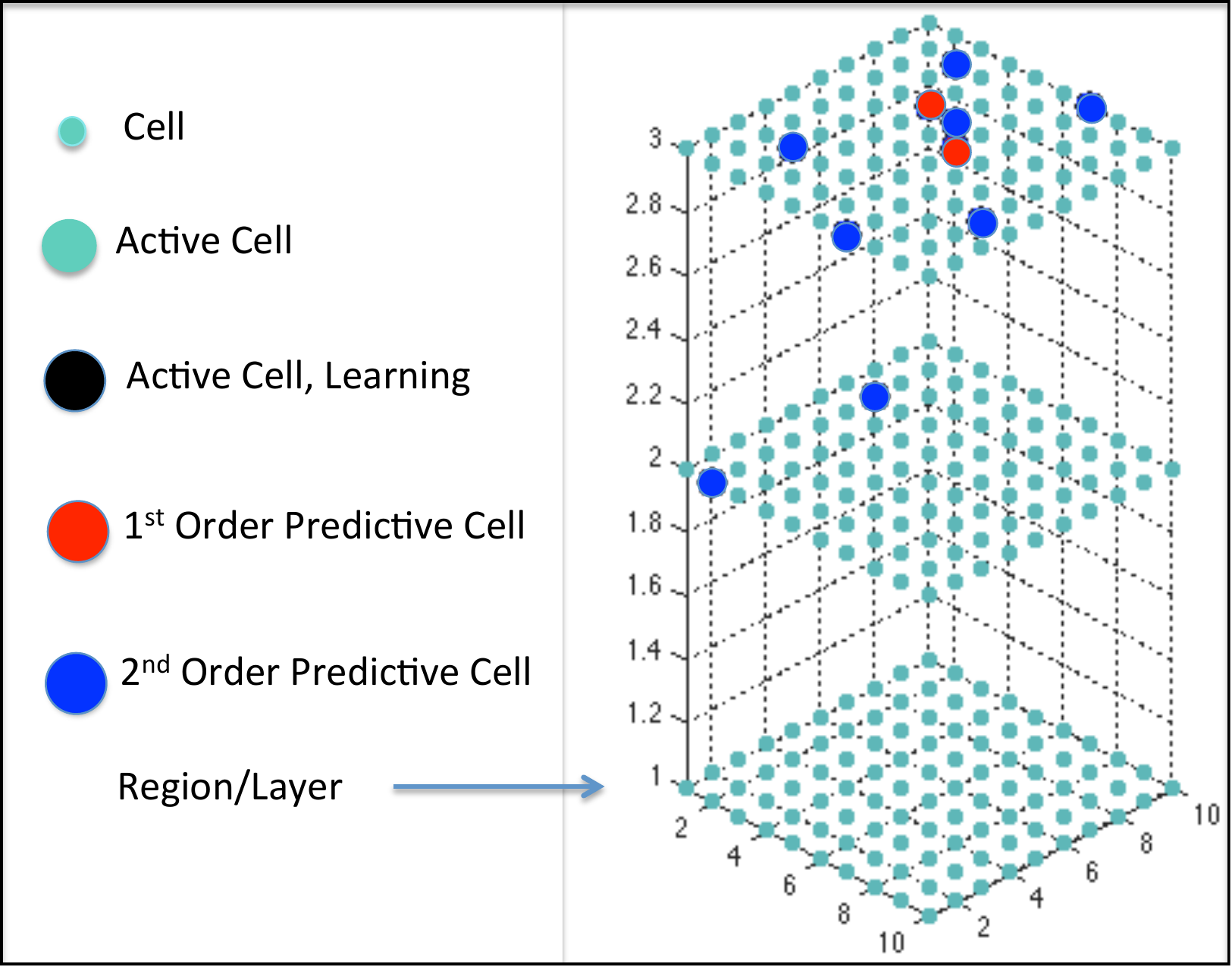}}
\caption{The prediction of number 2 using TM at different points in time. This number is presented to the network as a part of a 5-digit sequence. (a) Data is presented to the network. (b) First-order prediction. (c) Second-order prediction. (d) Cell color code. }
\label{predictionx}
\end{figure*}

\subsection {Prediction}
The HTM region has 3 cells per column. This gives the network the capability to predict first and higher order sequences. In this work, only the first-order and second-order sequences are verified.~\fig{predictionx} (a, b, and c) shows the prediction steps of number 2 in a synthetic sequence generated from a subset captured from EUNF dataset. Each subplot displays the number fetched to the network and a region of HTM that has 100 columns of 3 cells. In this context, subplots (a), (b), and (c) show the number presented to the network on the left and its representation at column and cell levels on the right. In this test, only 10 columns are used to represent the feedforward input in each time step. It is worth noting in~\fig{predictionx}-(a) that all the cells of the winning columns are active. This indicates that the network has never seen this particular input before and for this reason there is a \textit{bursting} state. The cells that colored in black will learn this particular input so that it can be predicted in the next time steps. After fetching the same sequence several times, the network can start predicting the next number.~\fig{predictionx}-(b) and (c) show the prediction of number 2 at time interval t+1 and t+2. 

\subsubsection{Prediction Accuracy}
\begin{figure} [h!tb]
\begin{center}
\includegraphics[width=60mm, height=55mm]{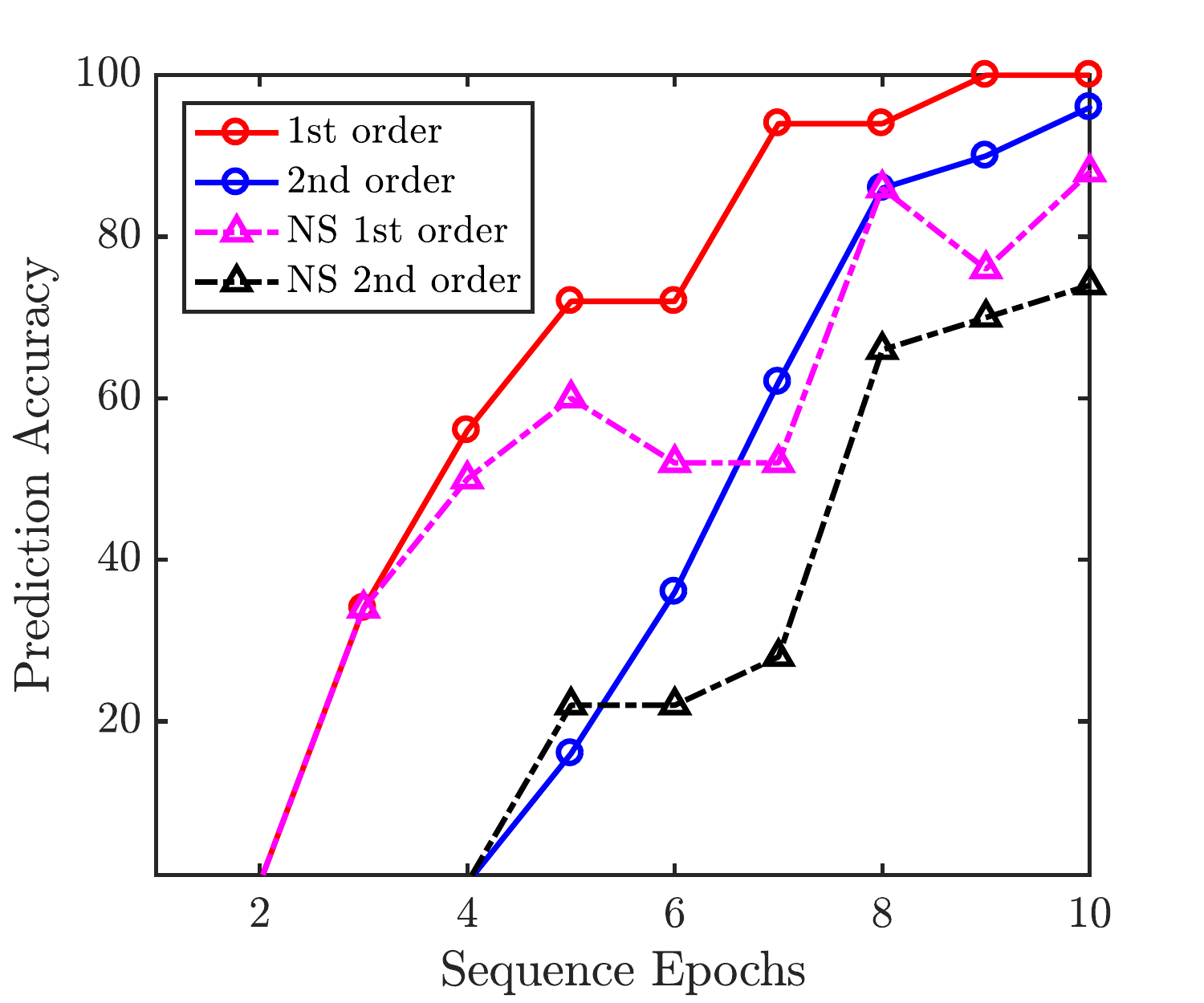}
\caption{First and second order HTM-TM prediction accuracy without and with the present of noise.}
\label{Pred_5}
\end{center}
\end{figure}

\fig{Pred_5} demonstrates the changes in the prediction accuracy of the HTM network for a 5- number long sequence. Notice that the prediction starts after fetching the same sequence a couple of times since synapses' permanence values at initial time step were 126 or 127. Once the segment has synapses with permanences over 128, it can be involved in the prediction step. This is to avoid  random predictions.~\fig{Pred_5}  also illustrates that the second order prediction starts after fetching the sequence four times, because the cell should establish connections to predict the first order sequences then second-order sequences.

\subsubsection{Prediction Accuracy in the Presence of Noise}
The presence of noise has a relatively small effect on the prediction accuracy in HTM due to the robust representation of input data as SDR. In this work, $\approx 25\%$ of salt \& pepper noise is added to the same sequence used in the previous test. ~\fig{Pred_5} explains the variation in the prediction accuracy in the presence of noise. It is obvious that the noise causes a reduction in accuracy, but this is still relatively small since it is only $\approx 12\%$ and $\approx 20\%$ for the first and second order predictions, respectively.

\subsection{Performance Analysis}
The proposed architecture is synthesized on Xilinx ZYNQ-7 FPGA, at 100 MHz. The vectorized MATLAB version runs on 2.9GHz Intel i7, 8-core MacBook Pro. We found that training SP on FPGA takes only 5.75 $\mu$s/sample, while the MATLAB model takes 6.893 $m$s/sample. In case of TM, we found that the training requires only 5.052 $\mu$s/sample. Since the proposed HTM architecture runs the SP and TM phases in a pipelined fashion, after the first iteration, the SP training time can be considered as the time required to process an input by the HTM network. The overall computation time of the SP is given in \eq{HTMCompTime} and \eq{HTMCompTimee}, which estimates the SP training time and how it is affected by the number of columns ($n_c$). The equations project the time needed to store the input packets in the MCU ($T_{StoreInput}$), the time to compute the overlaps and updates the synapses' permanences ($T_{OverlapComp.}$), and the time to move the data back and forth between the columns and the MCU ($T_{DataProp}$).

\begin{equation}
\begin{aligned}
\label{HTMCompTime}
 T_{SPCompTime}=&T_{StoreInput} + T_{OverlapComp.}\\
 &+T_{DataProp}
\end{aligned}
\end{equation}

\begin{equation} \label{HTMCompTimee}
\begin{aligned}
 T_{Training} (\mu{s})=&  \frac{1}{30} \times nc + T_{SPCompTime}
\end{aligned}
\end{equation}

\begin{table*}[h!tb]
\caption{Comparison of proposed HTM architecture with other hardware implementations on different ASIC and FPGA platforms. It is important to note that the implementations are on different substrates and this table offers a high-level reference template for HTM hardware rather than absolute comparison. Data is obtained from \cite{streat2016non, kerner2017hierarchical, james2017htm, li2016hardware}. }
\label{HardwareAnalysis}
\setlength\tabcolsep{3 pt}
\begin{center}
\begin{threeparttable}
\begin{tabular}{|c|c|c|c|c|c|}
\hline                      
\rowcolor{Gray} \textbf{Algorithm} & \textbf{Non-volatile HTM \cite{streat2016non}} & \textbf{Memrsitive HTM \cite{james2017htm}} & \textbf{LFSR HTM \cite{kerner2017hierarchical}} & \textbf{Digital HTM~\cite{li2016hardware}} & \textbf{This work (PIM HTM)} \\ \hline
  Task      & Classification   & Classification    & - & Prediction   & Classification \& Prediction  \\ \hline
  Operating Frequency   & - & - & 100 MHz &  100 MHz  & 100 MHz\\ \hline
  Proximal Segment Size & - & 9 & 64 & 1 & 16 \\ \hline
  Distal Segment Size   & - & - & - & - & 10 \\ \hline
  Device                & ASIC   & ASIC     & FPGA    & ASIC    & ASIC \\ \hline
  Power consumed per cell        & x\tnotex{tnote:robots-a2} & x\tnotex{tnote:robots-a2}  & - & 0.52 mW\tnotex{tnote:robots-a1} & 1.39 mW \\ \hline
  Dataset & MNIST &	AR, YALE, ORL, \& UFI &	- & MNIST & MNIST \& EUNF\\ \hline
  No. of columns x cells	 & 49-784x1 & 9x1 & 	200x1 & 400x2 & 100x3 \\ \hline
  Speed-up     &   -   &  -      &     110x         &  329x-676x  & 1364x    \\ \hline 
  Technology node  &    TSMC 180 nm     &  IBM 180 nm      &     -         &  Nangate opencell 45 nm       & TSMC 65 nm     \\ \hline
  Pooling                & Spatial     & Spatial   & Spatial    & Spatial-temporal     & Spatial-Temporal \\ \hline
  \end{tabular}
      \begin{tablenotes}
      \item\label{tnote:robots-a1} In the Digital HTM~\cite{li2016hardware}, the power of the register files is not included.
      \item\label{tnote:robots-a2} In these references, only the power of the SP is reported. Therefore, it is not included in this table as it does not reflect the actual amount of power consumed by total HTM region.
    \end{tablenotes}
\end{threeparttable}
\end{center}
\end{table*}

\subsection{Device Resources Utilization}
The FPGA resources utilized to implement the HTM region is reported in~\tb{FPGAResource}. The FPGA xc7z045ffg900-2, has 545 block RAMs, 218,600 and 70,400 logic and memory slices, respectively. For each column with 3 cells, 630 logic slices are used. This can be added to the resources used by the MCU to get 63,449 logic slices as a total for the entire HTM network with 100 columns. Also, it can be noticed from the RTL diagram of each column and cell that there are memory banks used to store synapses' strengths, cells status, and other region information. These memory banks are mapped to single-port (S) and quad-port (M) memory slices of size: 128x1, 32x1, 16x1. As a result, for 100 columns of 3 cells, 9,108 memory slices are needed. This means that the network can be scaled up to 300 columns for the given FPGA.

\begin{table}[h!t]
\caption{FPGA resources utilization (Xilinx xc7z045ffg900-2) of a single HTM region of 100 columns (3 cells per column)}
\label{FPGAResource}
\begin{center}
\begin{tabular}{ | c | c  | c | c |} 
 \hline
 \rowcolor{Gray} \textbf{Unit} & \textbf{LUT as Memory} & \textbf{LUT as Logic} \\ \hline
  HTM Region (SP+TP) & 9,108 - 12.94\% &  63,449 - 29.03\%  \\ \hline
  Column (3 cells) & 91 - 0.13\% & 630 - 0.29\% \\ \hline
  Cell Unit & 82 - 0.12\% & 512 - 0.23\%  \\ \hline
  MCU   &  8 - 0.01\% & 3609 - 1.65\%      \\ \hline
  
  \hline  
\end{tabular}
\end{center}
\end{table}

\subsection{Power Consumption}
The proposed design is synthesized in Synopsys using TSMC 65nm technology node while running the system at a 100 MHz clock. The power consumption of a single cell within a column is $\approx 1.39 mW $. Due to lack of full-scale architecture design and implementation of the spatial and temporal memory of an HTM in the literature, this work is compared with a closely relevant design, shown in \tb{HardwareAnalysis}. In \cite{li2016hardware}, the power consumed by a single cell is measured to be 0.52 mW. However, this measurement does not account for the power consumed by the cell register file (memory). In \cite{streat2016non,james2017htm} only the power of the spatial pooler is reported, which is estimated to be 5.17 mW and 31.56 $\mu$W, respectively. However, these numbers do not reflect the actual amount of power consumed by HTM region. 

\section{Conclusions}
The main contributions of this work are the development of a scalable architecture for the complete HTM algorithm, demonstrated on multiple machine-learning tasks. This work introduces synthetic synapses to overcome the hardware constraints such as  potential synapses. The proposed architecture when realized on Xilinx ZYNQ-7 platform shows a training speed up of $\approx$ 1364X over the software counterpart. Furthermore, the simulation results demonstrate that the spatial pooling with a large number of winning columns and small overlap threshold can improve the classification accuracy. The spatial pooler can handle $\approx 20\%$ of noise without affecting its performance significantly. On the other hand, the temporal memory is verified for sequence prediction task. It is shown that the proposed HTM hardware can perform first and second order predictions efficiently even in the presence of noise.

\section*{Acknowledgment}
The authors would like to acknowledge Jeff Hawkins and Subutai Ahmed from Numenta in clarifying the HTM theory; Seagate for sponsoring part of this research; the reviewers for their time and extensive feedback to enhance the quality of the paper.


\begin{IEEEbiography}[{\includegraphics[width=1in,height=1.25in,clip,keepaspectratio]{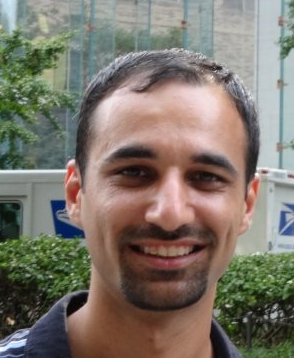}}]{Abdullah M. Zyarah} is a lecturer at the Department of Electrical Engineering, University of Baghdad. He specializes in digital and mixed signal designs and his current research interests include Neuromorphic architectures for energy constrained platforms and biologically inspired algorithms. Mr. Zyarah received his B.Sc. degree in Electrical Engineering from University of Baghdad, Iraq, in 2009, and M.Sc. degree in the same discipline from Rochester Institute of Technology, USA, in 2015. Currently, he is pursuing his PhD degree within the  Neuromorphic AI Lab research group in the Department of Computer Engineering, Rochester Institute of Technology. Prior to that, Mr. Zyarah worked as a lab instructor, from 2009 until 2013, and as a lecturer, from 2015 until 2017, at the Department of Electrical Engineering, University of Baghdad. 
\end{IEEEbiography}

\begin{IEEEbiography}
[{\includegraphics[width=1.25in,height=1.2in,clip,keepaspectratio]{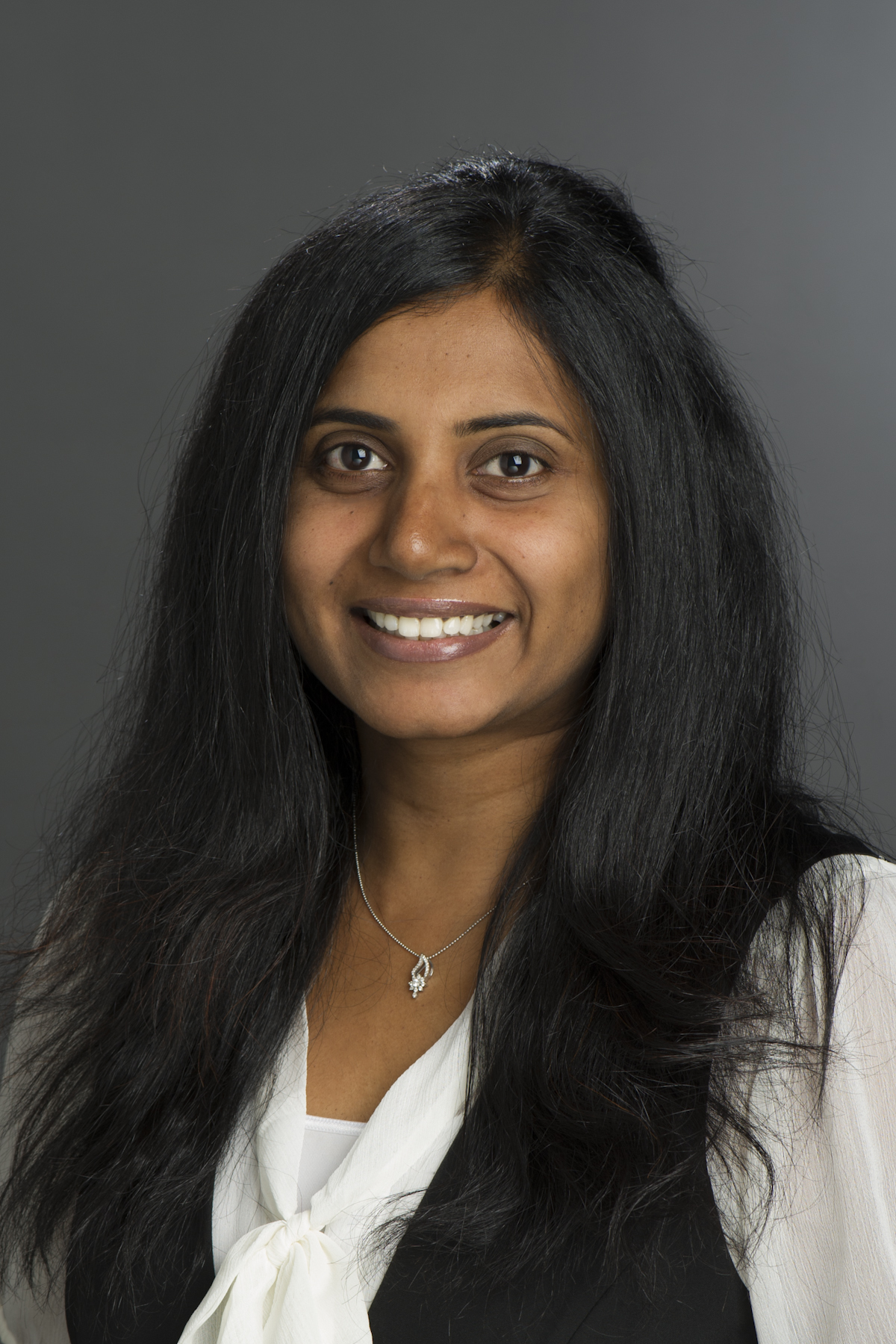}}]{Dr. Dhireesha Kudithipudi} is a professor and Director of the Neuromorphic AI lab.  Her research interests include brain-inspired computing, neuromorphic algorithms and hardware, emerging technologies for Machine Intelligence, and energy-efficient computing. She is also serving as the director for the Center for Human-Aware AI at RIT. Kudithipudi is the associate editor for IEEE Transactions on Neural Networks and Learning Systems. She received a PhD in computer engineering from the University of Texas at San Antonio and MS in Computer Engineering from Wright State University. Contact her at dxkeec@rit.edu.
\end{IEEEbiography}
\end{document}